\documentclass[11pt]{article}

\usepackage[margin=1in]{geometry}
\usepackage{newtxtext,newtxmath}     
\usepackage[utf8]{inputenc}
\usepackage[T1]{fontenc}
\usepackage[american]{babel}
\usepackage{microtype}

\usepackage{amsmath}
\makeatletter
\@ifundefined{openbox}{}{}
\@ifundefined{Bbbk}{}{}
\makeatother
\usepackage{amsthm}
\usepackage{amssymb}
\usepackage{graphicx}
\graphicspath{{plots/}}
\DeclareGraphicsExtensions{.pdf,.png,.jpg}

\usepackage{subcaption}
\usepackage{booktabs}
\usepackage[ruled,vlined,linesnumbered]{algorithm2e}
\usepackage{float}
\usepackage{enumitem}
\usepackage{tikz}
\usetikzlibrary{arrows.meta,positioning,decorations.pathreplacing,calc}
\usepackage{multirow}
\usepackage{booktabs,array,tabularx}
\newcolumntype{Y}{>{\raggedright\arraybackslash}X} 
\usepackage[numbers,sort&compress]{natbib}
\setcitestyle{square}

\usepackage{hyperref}

\def\betaMean{0.5} % 0.496
\def\betaCI{0.02} % 0.019
\def\rTwoMean{0.98} % 0.982
\def\rTwoCI{0.005}
\def\NSeeds{7}

\SetKwInput{KwInput}{Input}
\SetKwInput{KwOutput}{Output}
\SetKwInput{KwPar}{Parameters}
\SetKw{KwRet}{return}
\SetKwComment{tcp}{\ensuremath{\triangleright}\,}{}
\DontPrintSemicolon

\newcommand{\code}[1]{\ifmmode\text{\texttt{#1}}\else\texttt{#1}\fi}
\newcommand{\pmci}[2]{#1\,{\small$\pm$}\,#2}

\newcommand{\logodds}[1]{\log\!\frac{1-#1}{#1}}

\newcommand{\clip}[1]{\mathrm{clip}\!\left(#1\right)} % thin space, no visible "!"

\DeclareMathOperator{\TopK}{TopK}

\newif\ifshowarchive
\showarchivefalse % Set to true to show archived figures

\title{Budgeted Broadcast: An Activity-Dependent Pruning Rule for Neural Network Efficiency}
\author{%
\parbox{\textwidth}{\centering
Yaron Meirovitch$^{1,*,\dagger}$\quad
Fuming Yang$^{1,*,\dagger}$\quad
Jeff W.\ Lichtman$^{1,\dagger}$\quad
Nir Shavit$^{2,3,\dagger}$\\[3pt]
$^{1}$Harvard University\quad
$^{2}$MIT\quad
$^{3}$Red Hat AI\\[6pt]
\small $^{*}$Equal contribution\\ $^{\dagger}$Corresponding author: \texttt{yaron.mr@gmail.com, fumingyang@fas.harvard.edu, \\jeff@mcb.harvard.edu, shanir@mit.edu}
}%
}
\date{} 

\begin{document}
\maketitle
% ==========================================================
% Abstract (single paragraph)
% ==========================================================
\begin{abstract}
%%% old version
%Current pruning methods rank network components by their contribution to the loss. We introduce the orthogonal principle of \emph{metabolic cost}, inspired by resource constraints in biological circuits. We formalize this cost as a neuron's `traffic,' the product of its long-term activity ($a_i$) and its fan-out ($k_i$). Our rule, \textit{Budgeted Broadcast (BB)}, prunes a unit's weakest connections only when its traffic exceeds a budget. In sharp contrast to methods that punish inactive neurons, BB inherently protects highly selective, rare-feature detectors. This mechanism drives the system toward an emergent \emph{selectivity–audience balance}, $\log\!\tfrac{1-a_i}{a_i} \propto k_i$, a ``logit–fan-out'' law we derive and validate. Across diverse benchmarks, BB improves tail-event performance. On the challenging task of identifying synapses in brain imagery, it substantially outperforms standard pruning, achieving the highest F1-score and PR-AUC. The rule is simple and local, framing budget not as a constraint, but as a useful inductive bias for learning efficient representations.
Most pruning methods remove parameters ranked by impact on loss (e.g., magnitude or gradient). We propose Budgeted Broadcast (BB), which gives each unit a local traffic budget—the product of its long-term on-rate $a_i$ and fan-out $k_i$. A constrained-entropy analysis shows that maximizing coding entropy under a global traffic budget yields a selectivity–audience balance, $\log\!\tfrac{1-a_i}{a_i}=\beta k_i$. BB enforces this balance with simple local actuators that prune either fan-in (to lower activity) or fan-out (to reduce broadcast). In practice, BB increases coding entropy and decorrelation and improves accuracy at matched sparsity across Transformers for ASR, ResNets for face identification, and 3D U-Nets for synapse prediction, sometimes exceeding dense baselines. On electron microscopy images, it attains state-of-the-art F1 and PR-AUC under our evaluation protocol. BB is easy to integrate and suggests a path towards learning more diverse and efficient representations.
\end{abstract}

% ==========================================================
% 1 Introduction
% ==========================================================
\section{Introduction}\label{sec:intro}

Biological neural circuits are masterpieces of efficiency, sculpted by evolution to operate under strict metabolic and material constraints. This constant pressure for resource optimization fosters diverse and robust neural codes capable of navigating a complex world. In stark contrast, modern deep neural networks, trained with abundant compute, often learn highly redundant representations and falter on rare, long-tail events. This gap raises a central question: can we instill a formal principle of biological resource efficiency into artificial neural networks to make them more robust and diverse?

To date, the vast literature on network pruning has focused almost exclusively on a neuron's \textit{utility}: its importance as measured by weight magnitude, gradient information, or contribution to the loss. We argue that this network-level mechanism is overly opportunistic. Inspired by formal models of metabolic constraints in neuroscience \citep{barber1999}, we introduce the orthogonal axis of a neuron's \textit{cost}: the resources it consumes to broadcast its signal.

We formalize this cost as a neuron's \textit{traffic}, $t_i = a_i k_i$: the product of how often it `speaks' (its long-term firing rate, $a_i$) and the size of its `audience' (its axonal fan-out, $k_i$). Our method, \textbf{Budgeted Broadcast (BB)}, directly enforces a local budget on this traffic. In its simplest form, a unit prunes its weakest connections if and only if its traffic $t_i$ exceeds a threshold $\tau$. Intuitively, this has a direct consequence of protecting highly selective, rare-feature detectors (low $a_i$) by treating them as metabolically cheap, while curtailing the fan-out of over-active, low-selectivity units. This enforces a tradeoff: neurons can `speak' loudly to a small audience (high activity, low fan-out) or quietly to a large one (low activity, high fan-out), but not both. In contrast to ``lazy-neuron'' pruning (e.g., \citep{Hu2016NetworkTrimming}), BB reallocates connectivity toward a more efficient and diverse code.

\begin{figure*}[t]
\centering
\includegraphics[width=0.9\textwidth]{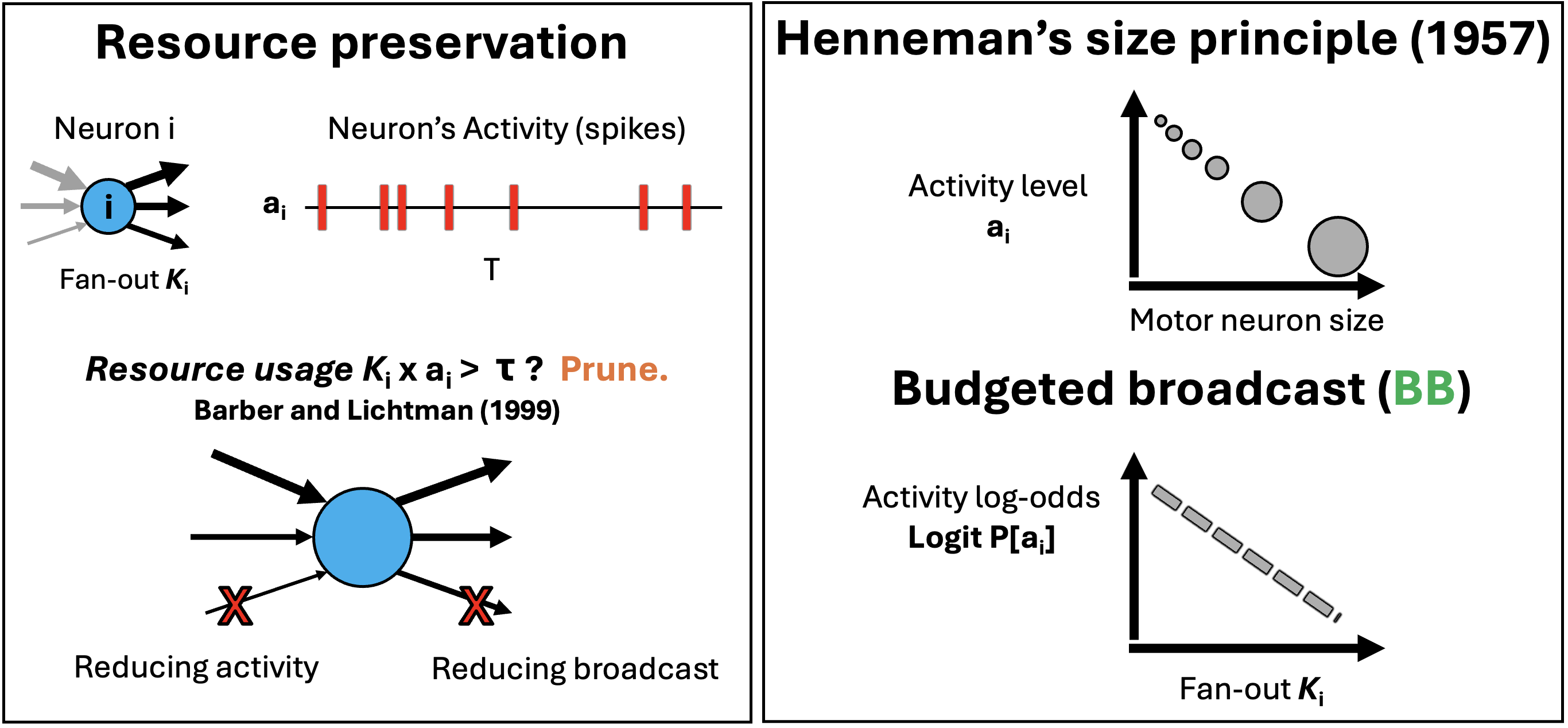}
\caption{\textbf{The conceptual framework of Budgeted Broadcast, from biology to a predictive theory.}
\textbf{(Left)} Our method models a neuron's metabolic cost as traffic, $t_i = a_i k_i$ (long-term activity $\times$ fan-out). If traffic exceeds a budget $\tau$, connections are pruned. This can be achieved by reducing fan-out (axonal pruning) or reducing fan-in to lower activity (dendritic pruning).
\textbf{(Top Right)} This rule is inspired by Henneman's size principle~\citep{Henneman1957SizePrinciple,Henneman1965SizePrinciple}, where large motor neurons (large size, analogous to fan-out $k_i$) have lower average activity levels ($a_i$).
\textbf{(Bottom Right)} Our resource-preservation rule predicts a linear relationship between a unit's fan-out ($k_i$) and its inactivity log-odds ($\log\frac{a_i}{1-a_i}$), which we term the selectivity-audience balance.
}
\label{fig:conceptual_overview}
\end{figure*}

This simple, local rule gives rise to a global organizing principle. An analysis of the network's coding entropy, which we detail later, predicts that this budget pressure drives the network to self-organize into a measurable equilibrium which we term \emph{selectivity–audience balance}  (Fig.~\ref{fig:conceptual_overview}, bottom right). 
%We show that non-saturated units naturally concentrate near a robust linear relationship between their fan-out and inactivity log-odds:
In learned codes where unit activities are only weakly correlated \citep{amari2002information}, this balance is attained when the unit's fan-out $k_i$ is proportional to its inactivity log-odds: 
\[
\log\frac{1-a_i}{a_i} \approx \beta\,k_i.
\]
This condition associates a unit's structure (node degree) to its function (node activity). We show that while it emerges as a regularity in a budgeted network, it is absent in networks trained (and/or pruned) with standard methods. In practice, we directly use this linear relationship to progressively modify the connections during learning. 
%The emergence of this predictable, law-like relationship validates that BB is not a mere heuristic, but a principled mechanism for enforcing resource efficiency.

% Conceptual map for the paper.

\paragraph{Contributions.}
Our contributions follow a  progression from empirical neuroscience to learning theory and ends with practical AI algorithms. First, we formalize a traffic budget, originally studied in the context of the neuromuscular connectome,  via a constrained-entropy objective, deriving a testable \emph{selectivity-audience balance} ($\logodds{a_i}=\beta k_i$) as the system's equilibrium state. We then implement this as a minimal, local controller, BB. We validate its properties on controlled didactic tasks, showing that it (1) verifiably produces the predicted balance, (2) provides structural safety for rare-but-relevant signals, and (3) overcomes optimization barriers that stall standard gradient-based methods. Finally, we demonstrate the utility of a general form of BB across four domains: ASR, face identification, synapse detection, and change detection, where it consistently improves tail/rare-event metrics at matched sparsity (Sec.~\ref{sec:asr}, \ref{sec:cv-vgg}, \ref{sec:cd}, \ref{sec:synapses}). 

% ==========================================================
% 2 Background \& Related Work
% ==========================================================

\section{Related Work}\label{sec:background}
Many pruning algorithms have been studied in the past decade. Magnitude pruning removes small weights and has underpinned compression work (\citealp{han2015learning,Han2015DeepCompression}); MorphNet uses layer-wise $L_1$ costs (\citealp{gordon2018morphnet}); Hessian and early saliency methods optimize local criteria (\citealp{lecun1990optimal}); the Lottery Ticket line studies sparse trainable subnets \citep{Frankle2018LotteryTicket}; and SynFlow prunes using a connectivity-sensitivity proxy \citep{Tanaka2020SynFlow}. Closest to our model is the bipartite-matching model of Dasgupta et al. \citep{Dasgupta2024Matching}, which simulates neural competition and reallocation of resources across outgoing edges.

Like Dasgupta et al., our approach draws inspiration from biological principles but differs fundamentally from existing pruning methods in both motivation and mechanism.

\textbf{Activity-dependent synapse elimination:}
Our work operationalizes a specific form of homeostatic regulation observed during neural development: activity-dependent synapse elimination. This process is captured by the two-force dynamic model of the neuromuscular junction of Barber and Lichtman \citep{barber1999}, in which a neuron's finite metabolic budget induces a trade-off between firing rate and audience size—high $a_i$ to few targets (low $k_i$) or low $a_i$ to many (high $k_i$). Our \emph{traffic} metric $t_i = a_i k_i$ is the direct computational expression of this trade-off. We translate the model's forces into our rule: (1) the \emph{presynaptic resource limit} becomes the budget gate $t_i>\tau$ that triggers pruning, and (2) \emph{postsynaptic competition} is modeled by removing the weakest outgoing weight $|w_{ij}|$. BB therefore implements structural homeostasis, turning foundational neurodevelopmental principles into a practical algorithm for sculpting network connectivity.

\textbf{Activity-Based Pruning:} Methods that prune based on activity ($a_i$) alone are an intuitive starting point, but they risk conflating a neuron's importance with its firing rate. %This approach does not readily distinguish a unit that fires usefully on common features from one that is non-selectively overactive.
In contrast, BB's traffic metric $t_i = a_i k_i$ is more nuanced in intuiting that a highly selective unit (low $a_i$) may be critically important and thus require a large audience (high $k_i$), hence protecting this 'quiet specialist.' % a role missed by activity-only criteria.

\textbf{Gradient-Based Methods:} SNIP and GraSP estimate importance from gradients~\citep{Lee2019SNIP,Wang2020GraSP}, while methods like RigL use gradient information to guide dynamic regrowth. While effective, these approaches rely on optimization signals that may lag optimal connectivity patterns. Unlike these gradient-driven methods, BB is a developmental controller derived from first principles. It operates using local, label-free statistics ($a_i, k_i$) and can reshape connectivity independently of gradient updates, acting as an autonomous homeostatic process analogous to biological circuit refinement.

\textbf{Structured Patterns:} While hardware-aligned patterns like N:M sparsity deliver predictable speedups, our focus is on the \emph{allocation principle} rather than the implementation pattern. BB can first allocate audience under a budget, then the resulting connectivity can be projected to hardware-friendly patterns for deployment-separating the biological principle from engineering constraints.

% ==========================================================
% 3 Method — Activity-Dependent Fan-Out Rule
% ==========================================================
\section{Method\,\textemdash\,Budgeted Broadcast (Local Broadcast Rule)}\label{sec:method}

%We now briefly describe the method (more details in {\bf Theory}). 
Our method, Budgeted Broadcast (BB), is governed by a local traffic-control rule. For each unit $i$, we periodically evaluate its traffic score:
\[ t_i = a_i \cdot k_i \]
where $a_i$ is the long-term average activation (on-rate), tracked via an Exponential Moving Average (EMA), and $k_i$ is its current fan-out. If $t_i$ exceeds a predefined budget $\tau$, the unit is marked for pruning in either or both ways: 1) A fraction of its weakest outgoing connections is removed (an `SP-out' action), directly reducing $k_i$ to bring the unit back within budget. 2) incoming connections are removed (an `SP-in' action) to reduce the neuron's activity $a_i$. These actions force a reallocation of network connectivity from high-traffic to low-traffic units.
In practice, we keep each unit's "audience" proportional to how quiet or busy it is. Let $\tilde a$ be a unit's activity Exponential Moving Average (EMA); the target degree is
\[ k \;=\; d_0 + \beta^{-1}\log\frac{1-\tilde a}{\tilde a}, \quad k\in[m,D]. \]
Every $\Delta$ step we recompute $k$ per unit and reselect Top-$k$ by $|W|$, enabling natural regrowth. We apply this at FFN fan-in (SP-in) and optionally fan-out (SP-out), with a variance-preserving rescale to keep layer scale stable.

\paragraph{Entropy maximization.}
This degree controller satisfies the conditions needed to globally maximize coding entropy $H(h)$ of the network, subject to a total traffic budget $\sum_i a_i k_i \le T_{\max}$. The Lagrangian  $\mathcal{L}=H(h)-\beta\bigl(\sum_i a_i k_i - T_{\max}\bigr)$ is stationary for $\log\frac{1-a_i}{a_i}=\beta k_i$ consistently with the controller (see Appendix for the full derivation).

\begin{figure}[t]
\centering
\includegraphics[width=1.0\linewidth]{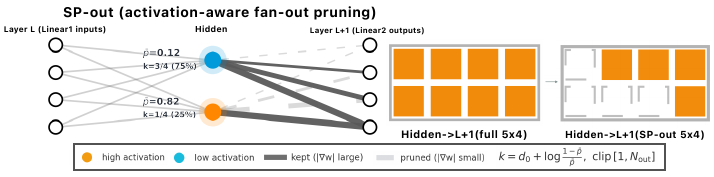}
\caption{\textbf{SP-out (Axonal pruning).} Activation-aware fan-out pruning that masks a hidden unit's outgoing connections to the next layer, enforcing the per-unit traffic budget $t=a\,k$ against a metabolic threshold $\tau$. High-activity units (large $a$) shed more outgoing edges; low-activity units keep more. Right: the learned binary mask sparsifies the dense hidden$\to L{+}1$ matrix according to $k=d_0+\tfrac{1}{\beta}\log\!\frac{1-a}{a}$, clipped to $[1,N_{\text{out}}]$. \textbf{SP-in} performs the complementary, opposite operation (fan-in pruning); see Appendix.}
\label{fig:sp-out}
\end{figure}

In practice, we implement BB inside FFN blocks (the $1\times1$ paths) by multiplying $W_1$ and $W_2$ with binary masks that refresh periodically (Fig.~\ref{fig:sp-out}). For simplicity, most of our theory is derived for the \emph{SP-out} actuators: at the first projection $W_1$, \emph{row masks} (SP-out@$W_1$) limit a source unit’s broadcast by reducing its fan-out $k$; at the second projection $W_2$, \emph{row masks} (SP-out@$W_2$) analogously limit a hidden unit’s broadcast. We provide in the appendix theoretical accounts for the complementary \emph{SP-in} actuator, implemented as \emph{column masks} at $W_1$ that reduce fan-in to modulate activity $a$ (Appendix). In this work, other components (e.g., attention, embeddings) remain dense. To minimize overhead, we avoid per-weight counters and store only a channel-wise EMA and the binary masks. %, which are refreshed via a simple Top-$K$ operation. 
While our method induces unstructured sparsity, mapping the learned masks to structured patterns (e.g., $N\!:\!M$ sparsity) is a deployment step deferred to future research.

\begin{algorithm}[t]
%\small
\caption{Budgeted Broadcast (SP-in/SP-out Refresh)}
\label{alg:bl}
\KwIn{$W\!\in\!\mathbb{R}^{R\times C}$; mask $M$; activity EMA $a_{\mathrm{ema}}$}
\KwOut{\unskip updated mask $M$}
\KwPar{schedule params: step, $T_{\mathrm{warmup}}$, $\Delta$;rule params: $\alpha,\beta,d_0$; bounds $m, D$; flag \code{SP\_IN}}\;
\If{\code{step} $<T_{\mathrm{warmup}}$ \textbf{or} \code{step} $\bmod \Delta \neq 0$}{\KwRet $M$}
\For{$u\leftarrow 1$ \KwTo $U$}{
  $k \leftarrow \clip\!\big(d_0+\beta^{-1}\log\tfrac{1-a_{\mathrm{ema}}[u]}{a_{\mathrm{ema}}[u]},\,m,\,D\big)$\;
  \uIf{\code{SP\_IN}}{$s \leftarrow \lvert W[:,u]\rvert$} \Else{$s \leftarrow \lvert W[u,:]\rvert$}
  $\mathrm{idx} \leftarrow \TopK(s,k)$ \tcp*[r]{\emph{Regrowth}: Top-$k$ from full row/column}
  \uIf{\code{SP\_IN}}{$M[:,u]\leftarrow 0$; $M[\mathrm{idx},u]\leftarrow 1$}
  \Else{$M[u,:]\leftarrow 0$; $M[u,\mathrm{idx}]\leftarrow 1$}
}
\KwRet{$M$}
\end{algorithm}

% ==========================================================
% 4 Theory
% ==========================================================
\section{Theory}\label{sec:theory}

A central question is why a simple, local pruning rule should lead to a globally coherent and efficient network structure. We get some insight by viewing our rule as a decentralized algorithm for solving a global optimization problem. Imagine we could design the network's connectivity to perfectly adhere to its function (a `god's-eye view') with the goal of maximizing the total information-coding capacity of the hidden units (measured by their entropy), subject to a fixed total `energy' budget.

While this constrained-entropy view implicitly leads to the selectivity-audience balance $\log\frac{1-a_i}{a_i}=\beta k_i$ (formally derived in the appendix), we can establish a more direct link between our local rule and the network's function using information theory. Under a standard noisy channel model for interlayer communication (see Assumption A1 in Appendix)
the mutual information $I(Z;Y)$ between a layer's code $Z$ and the next layer's preactivations $Y$ is upper-bounded by the trace of the output covariance:
$ I(Z;Y) \le \tfrac{1}{2\sigma^2}\operatorname{tr}(W^\top \operatorname{Cov}(Z) W). $
When correlations are weak (a regime BB and SGD promote and we observe empirically) and weights are bounded, this reduces to
$
I(Z;Y)\;\le\;\tfrac{C}{2\sigma^2}\sum_i a_i k_i,
$
so total traffic serves as a simple proxy for downstream information flow.(derivation in Appendix).
\[ I(Z;Y) \le \frac{C}{2\sigma^2}\sum_i a_i k_i \]
This indicates that the total traffic in a learning network serves as a tractable upper bound on the downstream information flow. Consequently, a BB refresh that prunes the weakest outgoing edges from high-traffic units produces a descent step on a composite objective $\mathcal{L} = \mathcal{L}_{\text{task}} + \lambda\sum_i a_i k_i$ (Lemma in Appendix). Hence, the observed network homeostasis observed in biological networks \citep{barber1999} and in our experiments is a consequences of optimizing a single, principled objective. Specifically, 
%a 5-seed Fashion-MNIST validation (Appendix Fig.~\ref{fig:theory_validation_multiseed}) shows 
we show that neurons in a budgeted network are more decorrelated than neurons trained with standard methods, while maintaining accuracy, and that total traffic is a good linear predictor of the estimated mutual information. 

% (conceptual)
\paragraph{Input versus output pruning.}
We also find that the two BB pruning actuators, SP-in and SP-out, provide complementary forces that drive the network toward this balance. A local linear-response analysis (see Appendix) shows that SP-in shocks primarily adjust a unit's activity ($a_i$), while SP-out shocks primarily adjust its audience ($k_i$). Together, the system can efficiently corrects deviations from the optimal state.

\begin{figure*}[t]
\centering
% ======================= FIGURE 2: Dual-Panel Balance Analysis =======================
% Data source: results/dual_panel_results.csv (Sept 12, 2024) - 7-seed validation
% Plot generation: plots/dual_panel_balance.pdf (Sept 12, 2024) via create_dual_panel_final.py
\includegraphics[width=1.0\linewidth]{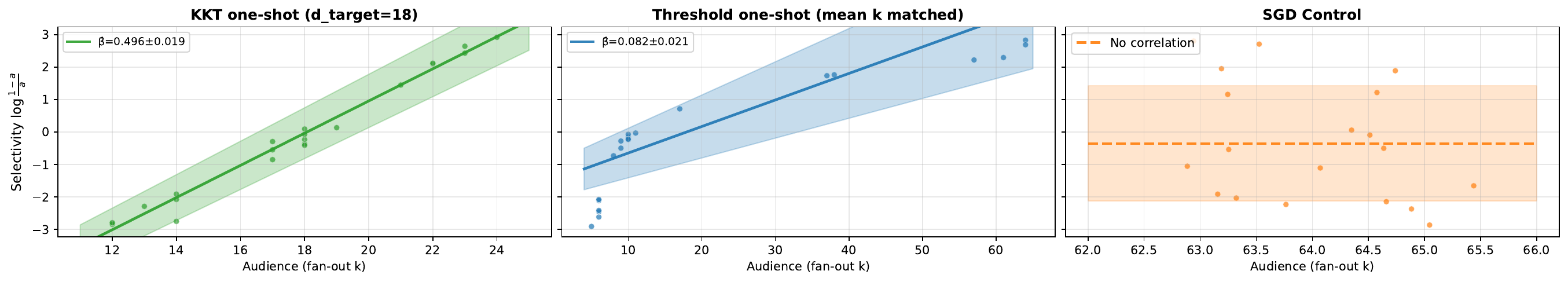}
\caption{\textbf{The selectivity–audience balance emerges under budget pressure on controlled XOR tasks.} The balance is a direct consequence of budget-driven structural adaptation, not an artifact of gradient-based training. \textbf{Left panel:} In networks trained with Budgeted Broadcast, a robust linear relationship %($R^2 = $\pmci{\rTwoMean}{\rTwoCI}, slope $\beta =$ \pmci{\betaMean}{\betaCI}) 
emerges between unit fan-out ($k_i$) and inactivity log-odds, confirming our theoretical prediction. 
%The representative scatter (seed 41) is overlaid with the mean fit line and confidence bands from \NSeeds{} independent seeds, demonstrating remarkable consistency. 
\textbf{Middle panel:} A one-shot traffic-threshold variant that prunes when $t_i = a_i k_i > \tau$ 
%(with $\tau$ calibrated per seed to match the KKT run’s mean fan-out) 
produces a similar trend but with a wider variability band and mild curvature, consistent with the threshold gate being a local approximation to the KKT stationary law $\log\frac{1-a_i}{a_i} = \beta k_i$.
\textbf{Right panel:} In control networks trained with SGD alone, fan-out remains constant at the initialization value (64), eliminating any correlation with activity (see Sec.~\ref{subsubsec:xor_balance})
%This dual-panel comparison demonstrates that the balance is a direct consequence of budget-driven structural adaptation, not an artifact of gradient-based training. Architecture: XOR 16$\to$32$\to$64$\to$1 (details in Sec.~\ref{subsubsec:xor_balance}).
}
\label{fig:balance}
\end{figure*}

% ==========================================================
% 5 Experiments
% ==========================================================
\section{Experiments}\label{sec:experiments}

%Our experiments are designed as a rigorous progression:
We first provide clean-room validation of BB's core properties on controlled didactic tasks (balance, safety for rare features, and overcoming optimization barriers), then demonstrate the principle's breadth on large-scale benchmarks (ASR, face identification, change detection), and conclude with a capstone test in synapse segmentation that exercises the method in a 3D U-Net for biomedical imaging.

\subsection{Didactic validation: mechanism, safety, and hardness}\label{subsec:didactic_validation}

We first use simple  MLP architectures to investigate three consequences of BB on controlled tasks—mechanism (XOR balance), feature safety (DNF+rare), and optimization hardness (DNF witness). %)—before evaluating applications on ASR and vision at matched compute.
%
%For our didactic experiments, 
While the specific controller implementation can vary (e.g., using a global budget with adaptive $\beta$ or a fixed local threshold $\tau$), all variants operate on the same core idea: pruning is triggered when a unit's traffic $t_i = a_i k_i$ becomes excessive. This allows us to cleanly study the emergence of the predicted balance, the inherent safety for rare features, and the ability to overcome optimization challenges.

\subsubsection{Emergence of the Selectivity-Audience Balance}\label{subsubsec:xor_balance}

To provide a visualization of the selectivity--audience balance, we use a simple 3-layer MLP trained on the XOR task (Input$\to$H1(64)$\to$H2(128)$\to$Output, with ReLU activations). We use SP-out on $W_2$ (row-mask on $W_2$) to control the output fan-out of the first hidden layer (H1). Activity ($a_i$) is measured as the post-ReLU EMA of the H1 units. As shown in Figure~\ref{fig:balance}, this setup produces a stable linear relationship between fan-out ($k_i$) and the log-odds of inactivity ($\logodds{a_i}$), ensuring that the BB mechanism achieves the theoretically predicted balance (100\% accuracy; linear fit with slope $\hat{\beta}=$\pmci{\betaMean}{\betaCI} and $R^2{=}$\pmci{\rTwoMean}{\rTwoCI} on non-saturated units across \NSeeds~seeds).

\subsubsection{DNF Tasks: Safety and Optimization}\label{subsubsec:dnf_witness}

We study two aspects of BB on Disjunctive Normal Form (DNF; an OR of several AND clauses) tasks: rare-feature safety and optimization barrier removal.

\begin{figure*}[t]
\centering
\begin{subfigure}[t]{0.32\textwidth}
  \centering
  \includegraphics[width=\linewidth]{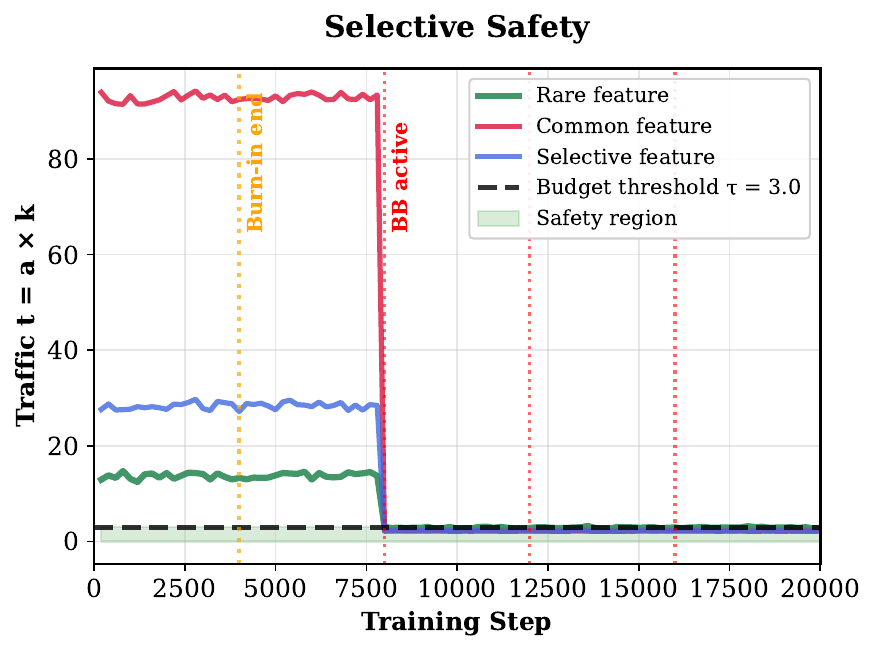}
  \caption{Selective Safety}
  \label{fig:didactic_safety}
\end{subfigure}\hfill
\begin{subfigure}[t]{0.32\textwidth}
  \centering
  \includegraphics[width=\linewidth]{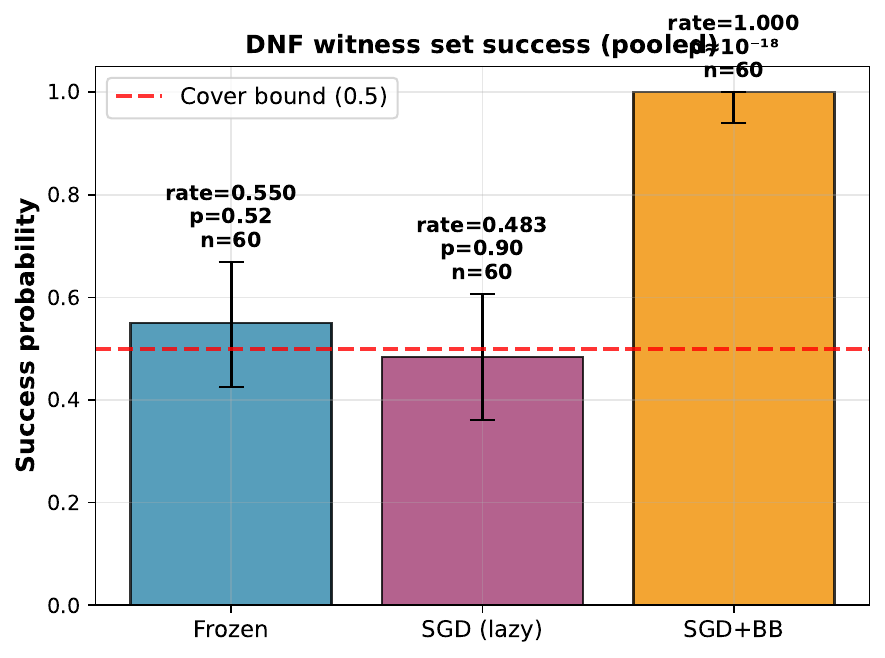}
  \caption{Optimization Barrier}
  \label{fig:didactic_success}
\end{subfigure}\hfill
\begin{subfigure}[t]{0.32\textwidth}
  \centering
  \includegraphics[width=\linewidth]{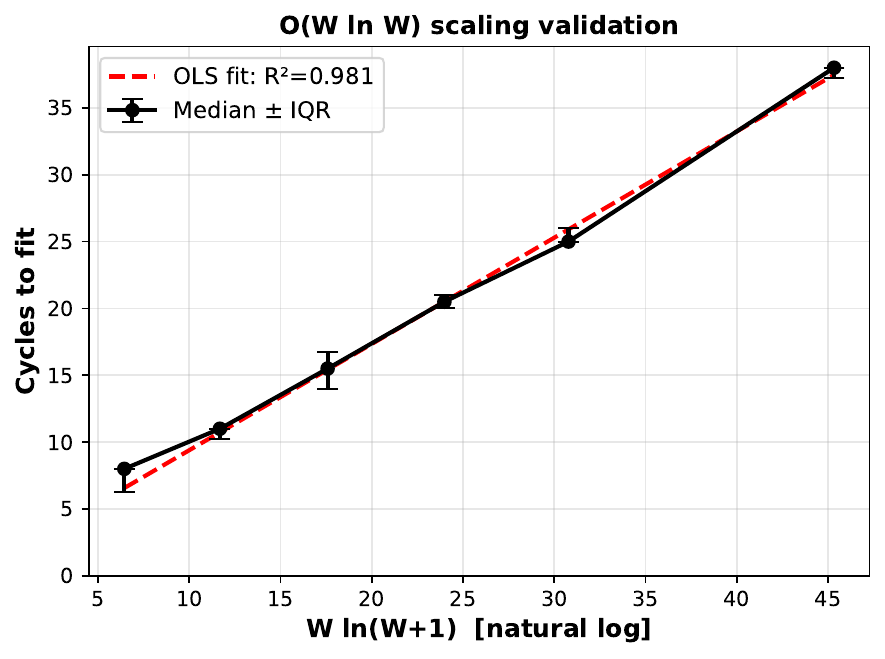}
  \caption{Convergence Scaling}
  \label{fig:didactic_scaling}
\end{subfigure}
\caption{\textbf{BB's core properties validated on controlled DNF tasks.} These experiments confirm the mechanism, safety, and optimization benefits of the BB principle.
\textbf{(a)} BB inherently protects rare features (green line), whose traffic remains safely below the budget $\tau$, while actively pruning over-active common features (red line).
\textbf{(b)} BB consistently solves a DNF task designed to make standard SGD fail, overcoming a lazy-learning barrier.
\textbf{(c)} The number of cycles for BB to solve the DNF task follows a predictable $O(W \log W)$ scaling law. All setup details are in Appendix.}
\label{fig:didactic_validation}
\end{figure*}

\textbf{Safety for Rare Features.} We first test if the BB rule is able to protect rare but important signals. We construct a DNF task containing features with varying frequencies of activation: rare ($p{\approx}0.11$), common ($p{\approx}0.72$), and moderately selective ($p{\approx}0.22$). As shown in Figure~\ref{fig:didactic_validation}a, the BB controller demonstrates remarkable selectivity. The rare feature's traffic ($t_s{=}a_s k_s$) is low and only moderately reduced to go below the pruning threshold $\tau$. In contrast, the common feature is actively managed, its traffic sharply curbed by pruning. This empirically validates that by budgeting traffic, BB can distinguish between features based on their usage patterns, safeguarding the pathways for infrequent events.

\textbf{Overcoming an Optimization Barrier.} To test BB's ability to reshape learning dynamics, we designed a DNF task that is difficult for standard gradient-based methods. The task uses $W+1$ disjoint clauses, where each AND clause operates on a unique set of inputs. The ideal network should learn a sparse ``one-unit-per-clause'' representation, allocating one hidden unit for each clause.

This setup creates a severe credit assignment problem for standard SGD, particularly in ``lazy'' learning regimes where weights change little from their random initialization. We train the network on a witness set, where each input is designed to activate only one specific clause. 
We predict that when a mini-batch contains witnesses for different clauses, the averaged gradient is weak and ambiguous, failing to specialize any single unit to its target clause, causing the network to get stuck (being unable to break the initial symmetry of its random weights). Theory predicts (and our experiments confirm) that such a learner will fail to solve the problem about half the time (Fig.~\ref{fig:didactic_validation}b),consistent with Cover's separability fraction (formalized in Appendix).

In contrast, alternating SGD with our BB controller consistently escapes this barrier. After a few SGD steps, units that responded non-specifically to multiple inputs develop slightly higher average activity. The BB controller, being agnostic to the ambiguous gradients, simply identifies these ``uselessly busy'' units by their high traffic and prunes their connections. This structural change breaks the learning symmetry, allowing other units to specialize and ``capture'' a clause in the next training phase. This iterative process acts as a powerful search mechanism. As shown in Figure~\ref{fig:didactic_validation}b and~\ref{fig:didactic_validation}c, BB consistently solves the task, and the number of cycles required scales predictably as $O(W \log W)$. This empirically matches the ``coupon collector'' behavior we formally analyze in the appendix), where the network ``collects'' the solution for each of the $W$ clauses one by one.

% If we need to reduce size we can move this to apptendix and refer to it in the Methods section where we mentioned Homeostatic properties
\paragraph{Homeostatic Resilience to Structural Shocks.}
Finally, we tested the dynamic resilience conferred by the BB rule. In a ``shock–recovery'' experiment, we subjected a trained network to sudden, large-scale pruning events and observed its response. The network exhibited graceful degradation in performance, followed by rapid, autonomous recovery once training resumed. This demonstrates that BB creates not just a statically efficient architecture, but a dynamically stable one with robust homeostatic properties. The full protocol and results are detailed in Appendix.

\subsection{Domain 1: Automatic Speech Recognition}\label{sec:asr}

%\paragraph{Experimental Setup.}
To test BB on a foundational sequence-to-sequence task, we employed a standard encoder-decoder Transformer trained on the LibriSpeech (\cite{Panayotov2015LibriSpeech}) \texttt{train-clean-100} dataset. For a controlled comparison, all methods (including baselines) followed an identical three-stage training schedule, beginning with decoder dense pre-training and encoder-only align training before enabling sparsification for the final full-transformer training.

To establish a fair and empirically-grounded sparsity budget, we applied the final network density of 0.85 for all baseline methods, and mask refreshes occurred every 25 optimizer steps with no regrowth rule (detail in Appendix). This setup allowed us to fairly evaluate the impact of different pruning principles on Word Error Rate (WER), particularly on rare words.
\begin{figure}[t]
  \centering
  \includegraphics[width=\linewidth]{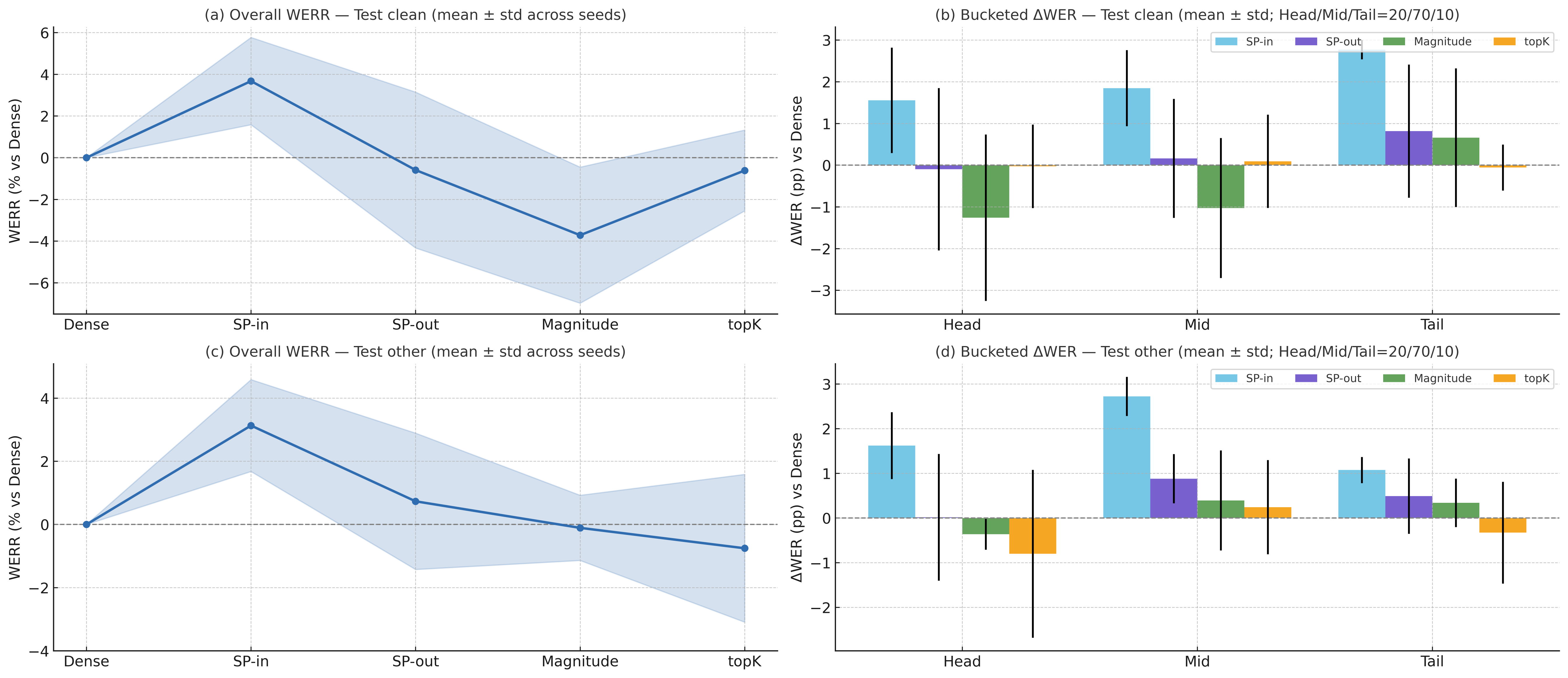}
%  \vspace{-2mm}
  \caption{\textbf{ASR on LibriSpeech.}
  (a) Overall Word Error Rate Reduction (WERR) \texttt{test\_clean};
  (b) Bucketed $\Delta$Word Error Rate (WER) \texttt{test\_clean} (Head/Mid/Tail fixed at 20/70/10; buckets are fixed across methods);
  (c) Overall WERR \texttt{test\_other};
  (d) Bucketed $\Delta$WER \texttt{test\_other}.
  Shaded bands/bars are mean~$\pm$~std over seeds; dashed line is Dense (WERR / $\Delta$WER$=0$).}
  \label{fig:asr-main}
\end{figure}

%\noindent\textbf{Overall results (WERR).}
Under the identical schedule and budget, BB (SP-in) is consistently best (Fig.~\ref{fig:asr-main}a,c), while BB (SP-out) is roughly neutral and Magnitude/Top-$k$ trails.

%\noindent\textbf{Bucketed results ($\Delta$WER).}
To localize gains, Fig.~\ref{fig:asr-main}b,d report \emph{bucketed} $\Delta$WER using the fixed Head/Mid/Tail buckets. We assign utterances to Head/Mid/Tail by sorting items by frequency and taking disjoint quantiles (20\%/70\%/10\%); buckets are fixed across methods and runs. All results are under matched budget, placement, schedule, and seeds. Averaged across seeds, SP-in improves all buckets and is largest on the long tail; SP-out shows smaller gains; Magnitude is negative on Head and near zero on Mid/Tail. This suggests that while magnitude pruning may harm performance on common words, BB's traffic-based approach reallocates resources to benefit the entire frequency spectrum, especially the challenging long tail.

\subsection{Domain 2: Face Identification}\label{sec:cv-vgg}

For face identification, we utilized a standard ResNet-101 (\cite{He2016ResNet}) backbone with its final layer adapted for the 7,001 identities in our curated VGGFace2-7k dataset \citep{Cao2018VGGFace2}. To test BB in a modern convolutional architecture, we applied it as a fan-in mask (SP-in) to the $1\times1$ projection kernels within each bottleneck block. This specific placement allows us to investigate the effect of budgeting traffic between channels in a ResNet. All sparse methods, including baselines like Magnitude pruning and RigL \citep{pmlr-v119-evci20a}, were applied to the same set of kernels to ensure a fair comparison based on Top-1 classification and verification accuracy.

We pre-specify the budgets before training. Concretely, we sweep six target sparsity levels 
\(s \in \{0.9, 0.7, 0.6, 0.5, 0.4, 0.3\}\)
and enforce the same target for all methods on the identical layer subset and fan-in masking side. Masks are refreshed every 200 optimizer steps with regrowth enabled at each refresh (i.e., previously pruned edges may re-enter via top-$k$). This protocol isolates the pruning principle itself under matched budgets and placement (details in Appendix).
\begin{figure*}[t]
  \centering
  \includegraphics[width=\linewidth]{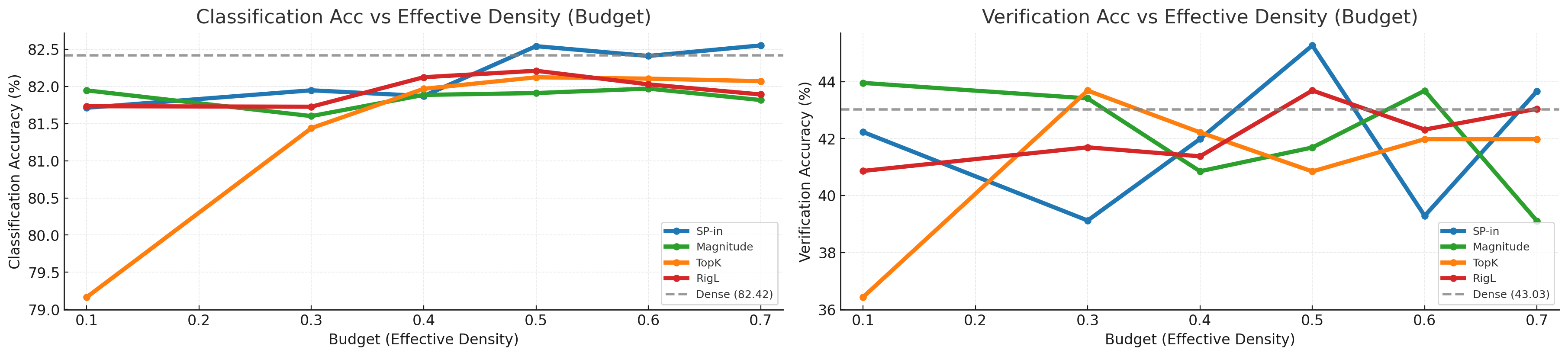}
 % \vspace{-2mm}
  \caption{\textbf{Pareto fronts on VGGFace2--7k.} \emph{Left}: Top-1 classification accuracy vs.\ budget (effective density). \emph{Right}: verification accuracy vs.\ budget on a held-out pair set. Each curve shows the best checkpoint per method at each density; the dense reference is the gray point at $1.0$. Across a broad range of budgets, SP-in forms or matches the upper envelope while using fewer active parameters}.
  \label{fig:cv-pareto}
\end{figure*}

For each density, we sweep 30\,epochs and pick the best validation checkpoint per method. Fig.~\ref{fig:cv-pareto} plots \emph{Top-1} (left) and \emph{verification} (right) against effective density. Across $0.3{\!-\!0.7}$, SP-in forms or matches the upper envelope and often exceeds the dense references around $0.5{\!-\!}0.7$. RigL is competitive at higher densities; magnitude degrades as sparsity increases; activation Top-$k$ shows inconsistent peaks but does not dominate.

\subsection{Domain 3: Change Detection}\label{sec:cd}

To evaluate BB's performance in a pixel-wise prediction task, we addressed bi-temporal building change detection on the LEVIR-CD dataset \citep{Chen2020}. We used a lightweight, Siamese encoder-decoder architecture (FC-Siam-conc) that processes two temporal images to produce a binary change mask. For this model, SP-in was applied as a fan-in mask to the first $3\times3$ convolution in each encoder block, with the decoder remaining dense. We report mean Intersection-over-Union (IoU) and F1-score on the held-out test set, comparing against the unpruned dense model under an identical training schedule. 

We compare BB(SP-in) against the dense model without pre-specifying a sparsity target using default hyperparameters. %and let the network learn its own level. 
This yields a final global density of 0.70. Masks use a warm-up of 1{,}000 optimizer steps, then refresh every 50 steps, with regrowth enabled at each refresh (i.e., previously pruned edges may re-enter via top-$k$). This protocol ensures a fair comparison under matched placement and schedule while allowing SP-in to discover an empirically grounded budget (details in Appendix).

\begin{figure*}[t]
  \centering
  \includegraphics[width=0.8\linewidth]{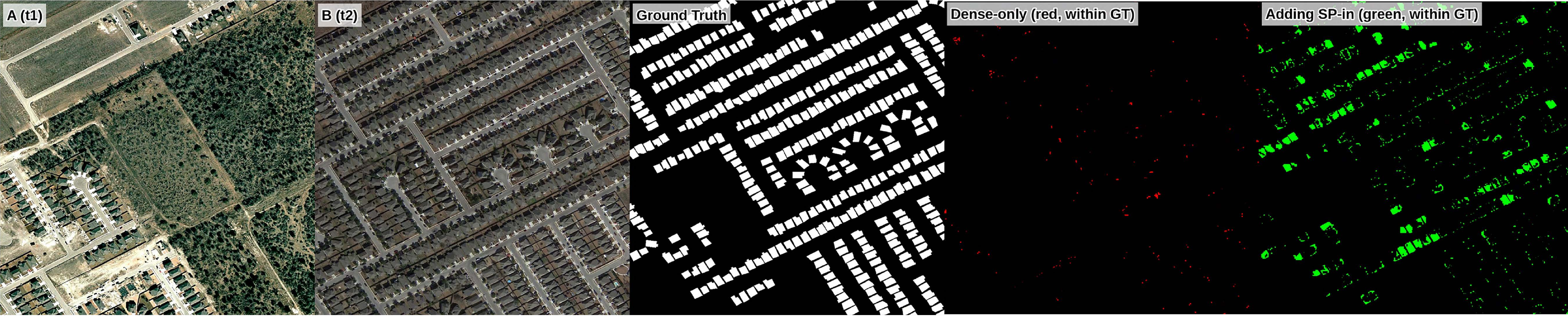}
 % \vspace{-2mm}
  \caption{\textbf{Change detection on LEVIR-CD: Dense vs.\ SP-in (top-$k$ qualitative).}
  Top: A (t1), B (t2), and Ground Truth. Bottom:
  \emph{Dense-only} TPs within GT (red), \emph{SP-in-only} TPs within GT (green)
  Shown is the test image with the highest $\Delta\mathrm{TP}$.}
  \label{fig:cd-qual-main}
\end{figure*}

Under the same 30-epoch schedule and fixed decision threshold, SP-in improves over Dense in all runs, as summarized below.

Averaged across runs, this represents a relative improvement of \textbf{+10.8\%} in IoU and \textbf{+7.9\%} in F1 (details in Appendix).

SP-in recovers substantially more true positives \emph{inside} the GT regions, especially for small, spatially scattered changes, while preserving major detections shared with Dense. 

% -------- Archived/Appendix-bound figures --------
\ifshowarchive
\begin{figure}[t]
\centering
\includegraphics[width=\textwidth]{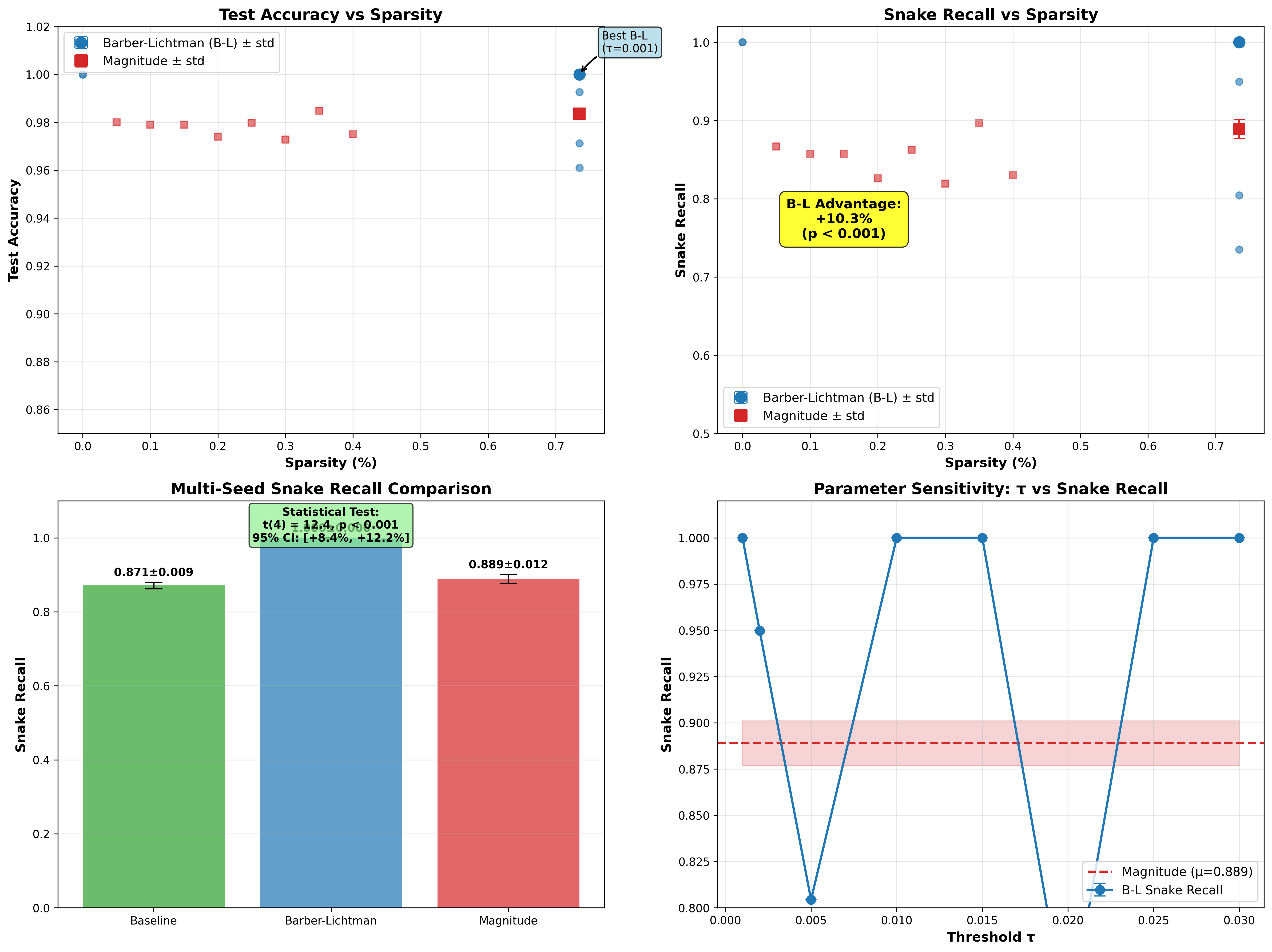}
\caption{\textbf{Sparsity--sensitivity frontier (archived for appendix).} Budgeted broadcast dominates magnitude pruning on rare-feature sensitivity at matched sparsity.}
\label{fig:pareto}
\end{figure}

\begin{figure}[t]
\centering
\includegraphics[width=\textwidth]{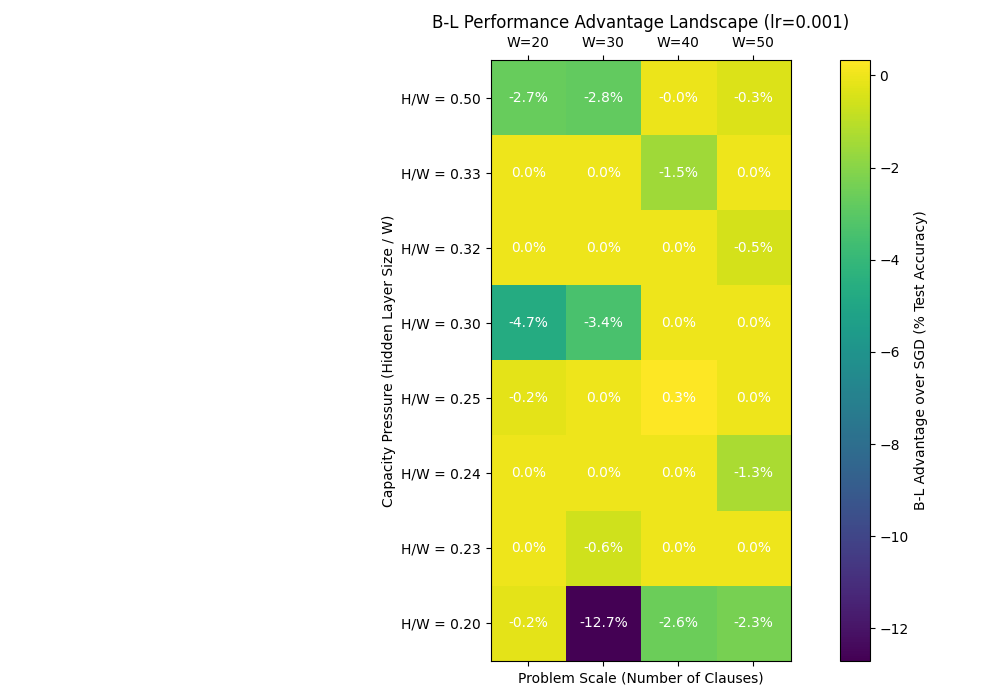}
\caption{\textbf{Performance landscape (archived).} Best seen under moderate capacity pressure ($H/W\approx0.25$).}
\label{fig:tract_heatmap}
\end{figure}

\begin{figure}[t]
\centering
\includegraphics[width=0.8\textwidth]{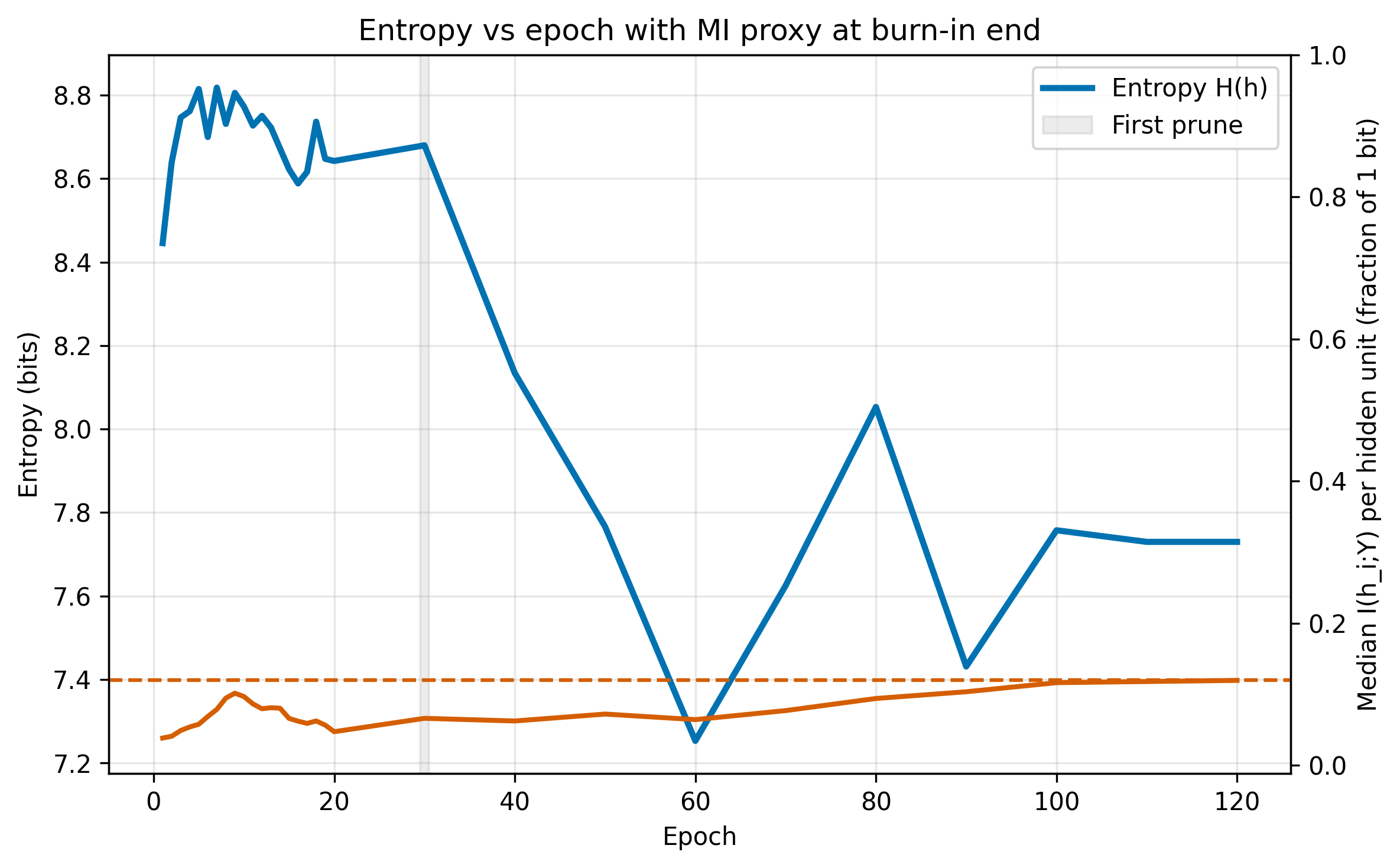}
\caption{MI trigger sanity check (archived): hidden-code mutual information vs percentile $\tau$ band.}
\label{fig:mi_vs_tau}
\end{figure}

\begin{figure}[t]
\centering
\includegraphics[width=\textwidth]{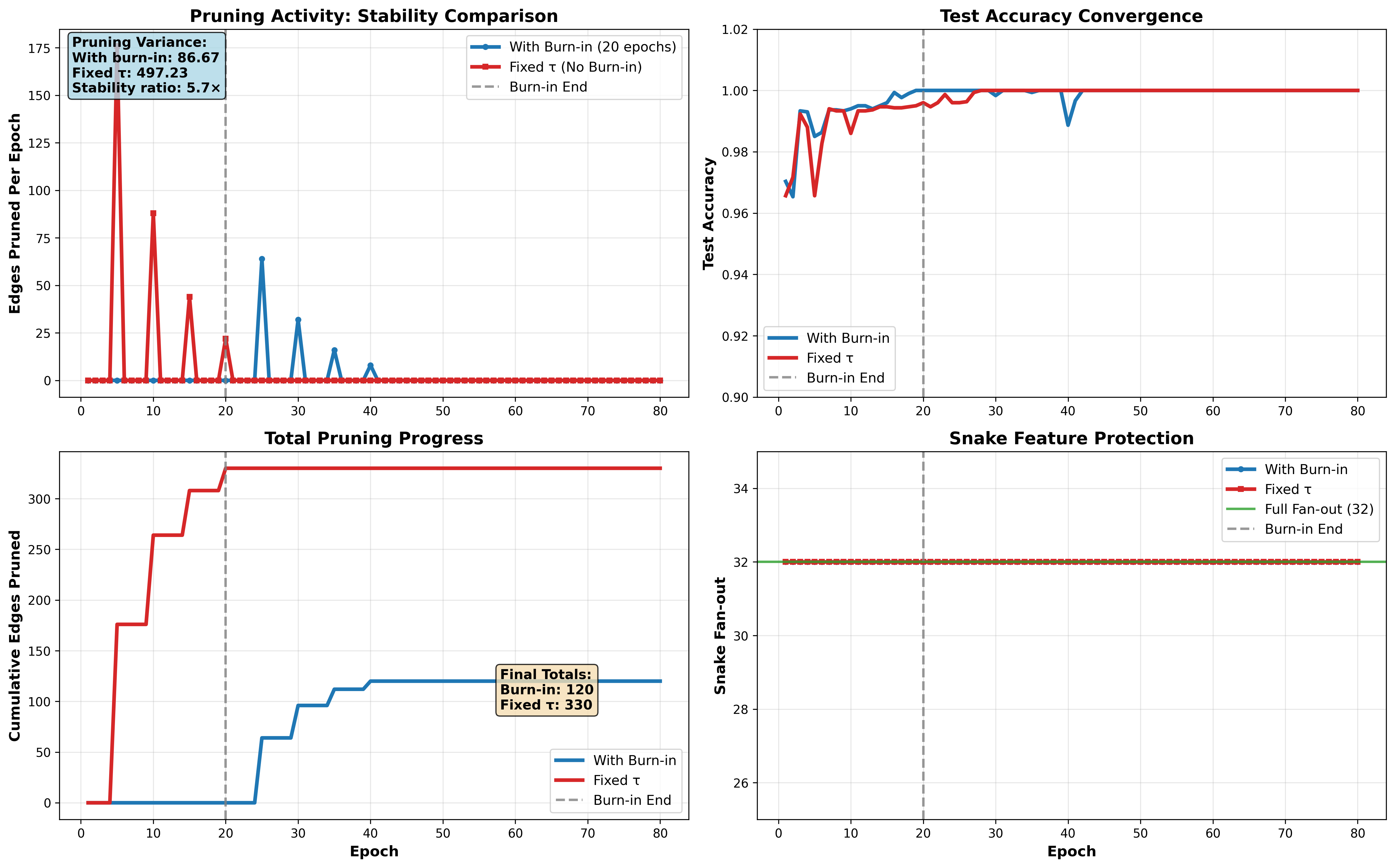}
\caption{Burn--in ablation (archived).}
\label{fig:burnin}
\end{figure}
\fi

% ==========================================================
% 5.x Synapse (moved up from after Discussion)
% ==========================================================
\subsection{Domain 4: Synapse Prediction (EM)}\label{sec:synapses}

As a capstone test of architectural generality, we applied BB(SP-in) (magnitude-based, row-wise fan-in masks) to a residual--SE 3D U-Net for synapse segmentation on volumetric EM from the SmartEM  dataset (\cite{Meirovitch2023SmartEM}; GT1 for training, GT2 held-out for testing). Concretely, we attach BB to all main $3\times3\times3$ convolutions (both conv1 and conv2) across encoder and decoder blocks, while leaving ConvTranspose upsampling layers and skip concatenations dense. We compare against a dense baseline and a standard magnitude pruning baseline, reporting PR-AUC and Best F1 on the held-out test set.

For synapse prediction, we use a fixed budget ratio of 0.70, apply a 1{,}000-step warm-up, then linearly ramp to the target over 8{,}000 steps; masks are refreshed every 200 optimizer steps, with variance-preserving rescaling $\sqrt{\text{prev}/\text{cur}}$ per output channel. Pruning is applied to all \texttt{Conv3d} layers in encoder and decoder blocks (including SE $1{\times}1{\times}1$ and residual $1{\times}1{\times}1$ projections), while \texttt{ConvTranspose3d} upsampling layers and skip concatenations remain dense. Dense and pruned models share the exact same pipeline; inference uses sliding windows with $8{\times}$ flip TTA, and we report PR-AUC and best F1 on the held-out GT2 set ((detail in Appendix)).

\begin{figure}[t]
  \centering
  % If available, show best-threshold vs shared-threshold qualitative panel
  \includegraphics[width=\linewidth]{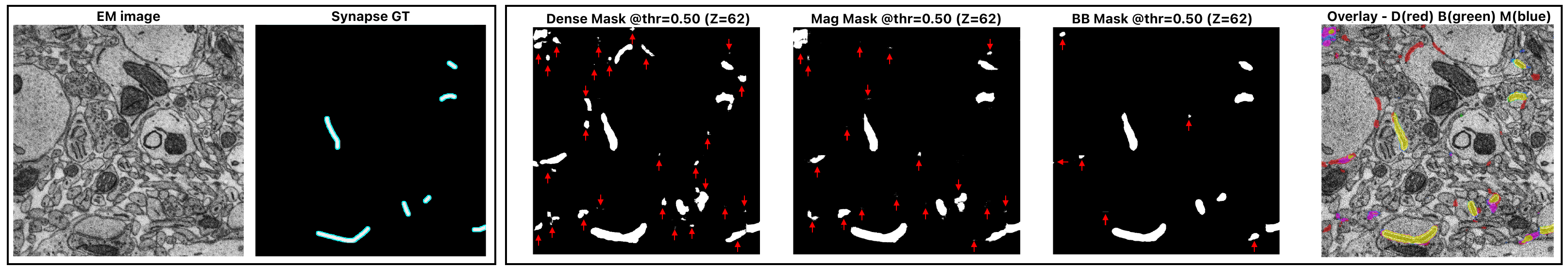}
  \caption{\textbf{Synapse prediction (per-method Best-F1).}
  Qualitative overlays and operating-point comparison. Red arrows denote false negatives
  (omitted GT synapses). 
  Right overlay: Dense=red, BB=green, Mag=blue; yellow marks consensus. Further details in Appendix.}
  \label{fig:synapse-qual}
\end{figure}

\begin{table}[t]
  \centering
  \small
  \setlength{\tabcolsep}{7pt}
  \begin{tabular}{lcccc}
    \toprule
    Method & PR-AUC & ROC-AUC & BestF1 & BestIoU \\
    \midrule
    Dense & $0.6952 \pm 0.010$ & $0.9889 \pm 0.0004$ & $0.6578 \pm 0.0070$ & $0.4906 \pm 0.0080$ \\
    BB (SP-in)    & $\mathbf{0.7407} \pm 0.014$ & $\mathbf{0.9906} \pm 0.0006$ & $\mathbf{0.6752} \pm 0.0090$ & $\mathbf{0.5099} \pm 0.0100$ \\
    Mag   & $0.7253 \pm 0.019$ & $0.9896 \pm 0.0009$ & $0.6643 \pm 0.0120$ & $0.4981 \pm 0.0140$ \\
    \bottomrule
  \end{tabular}
 % \vspace{-0.25em}
  \caption{\textbf{Synapse prediction (3 seeds, mean$\pm$std).} Results are computed at each method's own Best-F1 threshold and then averaged across seeds.}
  \label{tab:synapse-main}
\end{table}

\section{Discussion \& Future Work}\label{sec:discussion}

This work introduces a new axis for structural plasticity in artificial neural networks, shifting the focus from a component's \emph{utility} to its metabolic \emph{cost}. We formalized this cost as traffic ($a_i k_i$) and showed that a simple, local budget on this traffic can organize connectivity. The emergent selectivity–audience balance ($\log\frac{1-a_i}{a_i} \approx \beta k_i$) is a predictable equilibrium that links structure ($k_i$) to function ($a_i$). This computational framework provides a unified explanation for seemingly distinct biological phenomena from Henneman's size principle \citep{Henneman1957SizePrinciple} to the competitive dynamics of synapse elimination \citep{barber1999}, reframing them as convergent solutions to the universal problem of efficient information broadcast. The success of our Budgeted Broadcast rule on diverse benchmarks provides  empirical support for this structural perspective of neural organization.

Future work should study application of budgeted neural activity beyond FFNs and CNNs, and in particular to lateral connections and attention models. While our method introduces modest, amortized overhead from EMA tracking and periodic mask updates, its scalability makes it a promising candidate for foundation models where protecting the long tail of knowledge is paramount.
 
A Budgeted Attention mechanism would extend our per-neuron budget to a dynamic, per-token budget. A token's `traffic' could be defined as $t_j = f(A_j) \times k_{\text{eff}}(j)$, where $f(A_j)$ is a function of the token's activation norm (how `loud' it is) and $k_{\text{eff}}(j)$ is its \emph{effective fan-out}.

%% arXiv: bbl embedded

\bibliographystyle{unsrtnat}

\newpage
\part*{APPENDIX for Budgeted Broadcast: An Activity-Dependent Pruning Rule for Neural Network Efficiency}
Will be released with the published paper.

\end{document}